\documentclass[11pt,a4paper]{article}
\usepackage[hyperref]{acl2021}
\usepackage{times}
\usepackage{latexsym}

\usepackage{defs}

\usepackage{microtype}

\aclfinalcopy 
\def\aclpaperid{2433} 

\title{On Training Instance Selection for Few-Shot Neural Text Generation}

\author{ 
   Ernie Chang\thanks{ $\>$ Equal contribution. X.shen is now at Amazon Alexa AI.},  Xiaoyu Shen\footnotemark[1] ,  Hui-Syuan Yeh, Vera Demberg \\
   Dept. of Language Science and Technology, Saarland University
   \\
      {\tt \{cychang,xshen\}@coli.uni-saarland.de}
  \\ 
}

\date{}

\begin{document}
\maketitle
\begin{abstract}
Large-scale pretrained language models have led to dramatic improvements in text generation. Impressive performance can be achieved by finetuning only on a small number of instances (few-shot setting). Nonetheless, almost all previous work simply applies random sampling to select the few-shot training instances. Little to no attention has been paid to the selection strategies and how they would affect model performance. In this work, we present a study on training instance selection in few-shot neural text generation. 
The selection decision is made based only on the unlabeled data so as to identify the most worthwhile data points that should be annotated under some budget of labeling cost. 
Based on the intuition that the few-shot training instances should be diverse and representative of the entire data distribution, we propose a simple selection strategy with K-means clustering. 
We show that even with the naive clustering-based approach, the generation models consistently outperform random sampling on three text generation tasks: data-to-text generation, document summarization and question generation. 
The code and training data are made available at~\url{https://gitlab.com/erniecyc/few-selector}. We hope that this work will call for more attention on this largely unexplored area.
\end{abstract}

\section{Introduction}
Few-shot text generation is an important research topic since obtaining large-scale training data for each individual downstream task is prohibitively expensive. Recently, pretraining large neural networks with a language
modeling objective has led to significant improvement across different few-shot text generation tasks~\cite{ radford2019language, lewis-etal-2020-bart} and many techniques are proposed based on them~\cite{chen2019few,schick2020few,zhang2020pegasus,kale2020text,chang2020dart,chang2021neural,li2021prefix,chang2021jointly,chang2021does,su2020moviechats}. However, all the previous works simulate the few-shot scenario by randomly sampling a subset from the full training data. Little to no attention has been paid to the selection strategies. 
\begin{figure}[t]
  \centering
\includegraphics[width=\columnwidth]{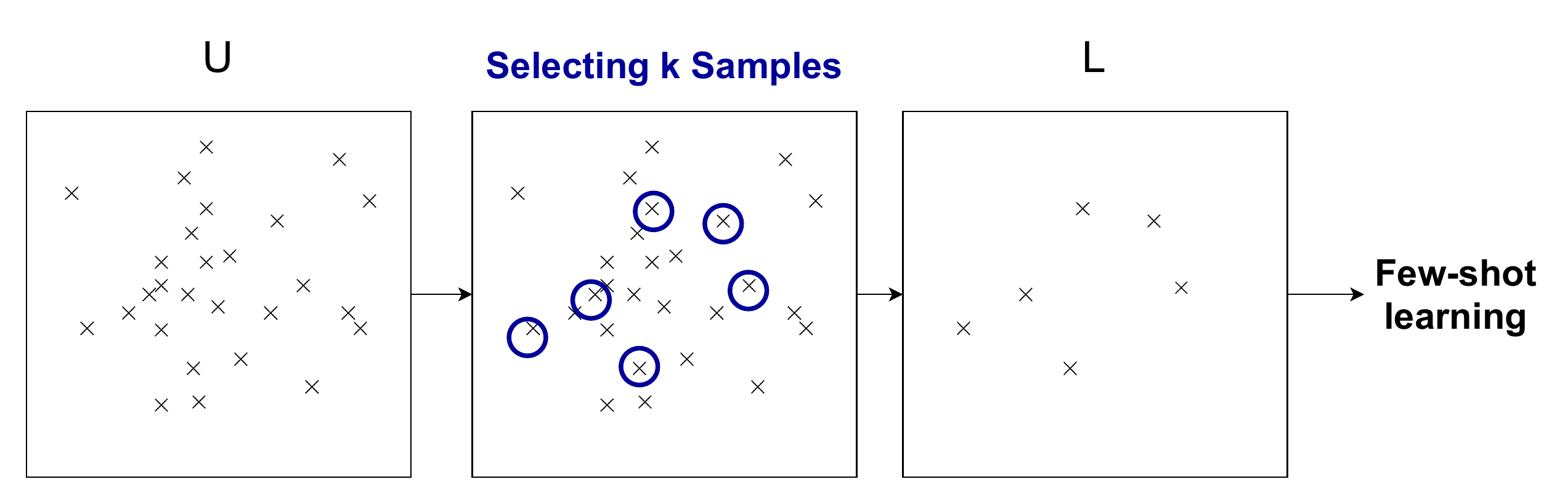}
\caption{ \small
\textbf{Training scenario}:  \textbf{U} represents unlabeled data and \textbf{L} indicates labeled instances. The annotation budget only allows selecting K data for annotating the reference text.
}
\label{fig:sample}
\end{figure}

In this work, we present a preliminary study at searching for an optimal strategy to select the few-shot training instances. Studying the selection strategy is motivated by two rationales. First, random sampling leads to a large variance of model performance~\cite{zhang2020pegasus,schick2020few,schick2020s}. Yet current works sample their own training data which makes it difficult to compare across different models. One can then not be sure whether an improved performance can be really ascribed to the model or the randomness of sampling. Using a stable selection strategy to find the most informative few-shot instances can provide a fair platform and better benchmark different few-shot generative models. Second, in practical applications, e.g.~document summarization, the training data is usually obtained by manually annotating the summaries for some selected documents. In Figure~\ref{fig:sample}, we illustrate the typical training scenario for text generation where the annotation budget only allows annotating a limited amount of data. Studying the optimal selection strategy can help make the most use of our annotation budget. Specifically, we focus on the label-free setting where \emph{the selection can only condition on the unannotated data}. Although leveraging the reference text may benefit the selection strategy, it conflicts with the realistic setting where we need to first select the data then get its annotated reference text.

The selection task resembles the theme of active learning~\cite{balcan2007margin}, where the model keeps identifying the most informative instances to get labeled. Existing active learning approaches can be roughly divided to uncertainty-based sampling and representative sampling~\cite{settles2009active}. Uncertainty-based sampling select samples
that maximally reduce the uncertainty of the model~\cite{tur2005combining}. This, however, requires a well-trained model with decent confidence score estimations in order to perform well. Therefore, in this paper, we opt for the representative-sampling where the selected training instances are expected to be dissimilar to each other and representative enough to cover all important patterns in the whole data distribution~\cite{agarwal2005geometric,wei2015submodularity}. This naturally matches the objectives of k-means clustering which minimizes the within-cluster variances while maximizing the between-cluster variances to encourage the diversity and representativeness of each cluster~\cite{krishna1999genetic,kanungo2002efficient}. As has been shown in image classification tasks, data points closer to the cluster centroids are usually most important, while other faraway points can even be safely removed without hurting model performance~\cite{kaushal2018learning,birodkar2019semantic}. Inspired by this, we propose a simple selection strategy which first clusters the whole unlabeled dataset with the K-means algorithm, and then from each cluster, selects the data point that is closest to the cluster centroid.

We conduct experiments on three popular text generation tasks: data-to-text, document summarization and question generation. The proposed selection strategy consistently outperforms random sampling and exhibits much smaller variance.
\paragraph{Contribution.} We present a preliminary study on training instance selection for few-shot text generation and propose a selection strategy based on K-means clustering. The proposed method shows consistent superior performance over random sampling, which can be used to make most use of the annotation budget in practical applications. Meanwhile, the selected training instances can serve as a better benchmark for few-shot text generation since they are not biased towards specific generative methods and do not have the large variance issue as found in random sampling. We further perform a set of ablation studies to analyze what contributes to a good selection. Our findings can also benefit research in active learning~\cite{konyushkova2017learning} since identifying the most informative training instances is a critical step before collecting more annotations through active learning.

\section{Problem Formulation}
Following the training scenario shown in Figure~\ref{fig:sample}, we denote the unlabeled data as ${U_1, U_2,\ldots, U_n}$ where n is the data size. Depending on the downstream task, ``data" can mean unlabeled structured data, documents and paragraphs respectively in the context of data-to-text, document summarization and question generation. We will select $K$ instances from the whole unlabeled dataset, annotate them with reference text, and then train a neural generative model based on the annotated data. $K$ is defined based on the annotation budget. In this work, since we focus on the few-shot scenario, $K$ is set to be small ($\leq 100$). The goal is to \emph{find the most representative $K$ instances that can lead to the optimal performance when trained on them}. 

\section{Selection by K-means Clustering}
The general idea of our proposed method is to first split the whole unlabeled data into $K$ clusters, then select one data point from each cluster. Specifically, we first map each data point into a vector, then cluster the vectors with the K-means algorithm. The objective is sum of the squared errors (SSE), which is also called cluster inertia:
\vspace{-0.2cm}
\begin{equation}
    SSE = \sum_{i=1}^{n}\sum_{j=1}^{K}w_{i,j}||x^{i}-\mu^{j}||_{2}^{2}
\end{equation}
where $\mu^{j}$ is the centroid of the $j$th cluster. $x^{i}$ is the embedding vector of $U_{i}$. $w_{i,j}=1$ if $x^i$ belongs to the cluster $j$ and $0$ otherwise. We optimize the objective function with the EM algorithm~\cite{dempster1977maximum} which iteratively assigns each data point into its closest cluster centroid. The initial centroid points are chosen based on the
K-means++ algorithm~\cite{arthur2007k}. The first cluster center is chosen uniformly at random from the data points, after which each subsequent cluster center is chosen from the remaining data points with probability proportional to its squared distance from the point's closest existing cluster center. By this means, we maximize the chance of spreading out the $K$ initial cluster centers. We use 10 random seeds for selecting initial centers and the clustering with the minimum SSE is chosen.

\begin{table*}
  \small
  \centering
  \resizebox{\textwidth}{!}{
  \begin{tabular}{lccc|ccc|ccc}
    \toprule
      & \multicolumn{3}{c}{ E2E } & \multicolumn{3}{c}{ CNNDM } & \multicolumn{3}{c}{ SQUAD } \\
    \cmidrule(r){2-4}
    \cmidrule(r){5-7}
    \cmidrule(r){8-10}
    \multirow{1}{*}  & 10 & 50 & 100 & 10 & 50 & 100 & 10 & 50 & 100 \\
    \cmidrule(r){2-4}
    \cmidrule(r){5-7}
    \cmidrule(r){8-10}
    Random
      &  4.38$\pm$7.12 &  11.57$\pm$4.29 &  26.22$\pm$2.58 
      &  13.51$\pm$6.47 &  24.81$\pm$3.77 &  35.24$\pm$2.89 
      &  1.23$\pm$6.22 &  3.33$\pm$5.89 &  7.65$\pm$3.61   \\      
    IC-Random
      &  2.15$\pm$4.58 &  9.80$\pm$2.62 &  24.71$\pm$2.71 
      &  12.30$\pm$3.89 &  24.71$\pm$2.45 &  33.29$\pm$1.92 
      &  1.34$\pm$3.23 &  1.79$\pm$3.77 &  6.97$\pm$2.55  \\   
    K-means
      &  \textbf{6.22}$\pm$2.33 &  \textbf{11.89}$\pm$1.39 &  \textbf{27.13}$\pm$2.22 
      &  \textbf{14.28}$\pm$\textbf{2.35} &  \textbf{25.19}$\pm$3.28 &  \textbf{36.31}$\pm$1.08 
      &  \textbf{1.56}$\pm$2.34 &  \textbf{4.77}$\pm$3.61 &  \textbf{9.33}$\pm$2.15  \\            
    \bottomrule
  \end{tabular}}  
\caption{\small Comparisons of random sampling, within-cluster random sampling (IC-Random) and K-means selection on the E2E, CNNDM, and SQUAD corpus (BLEU-4 reported).
}
\label{tb:benchmarks}
\end{table*}


\begin{figure}[t]
  \centering
\includegraphics[width=0.95\columnwidth]{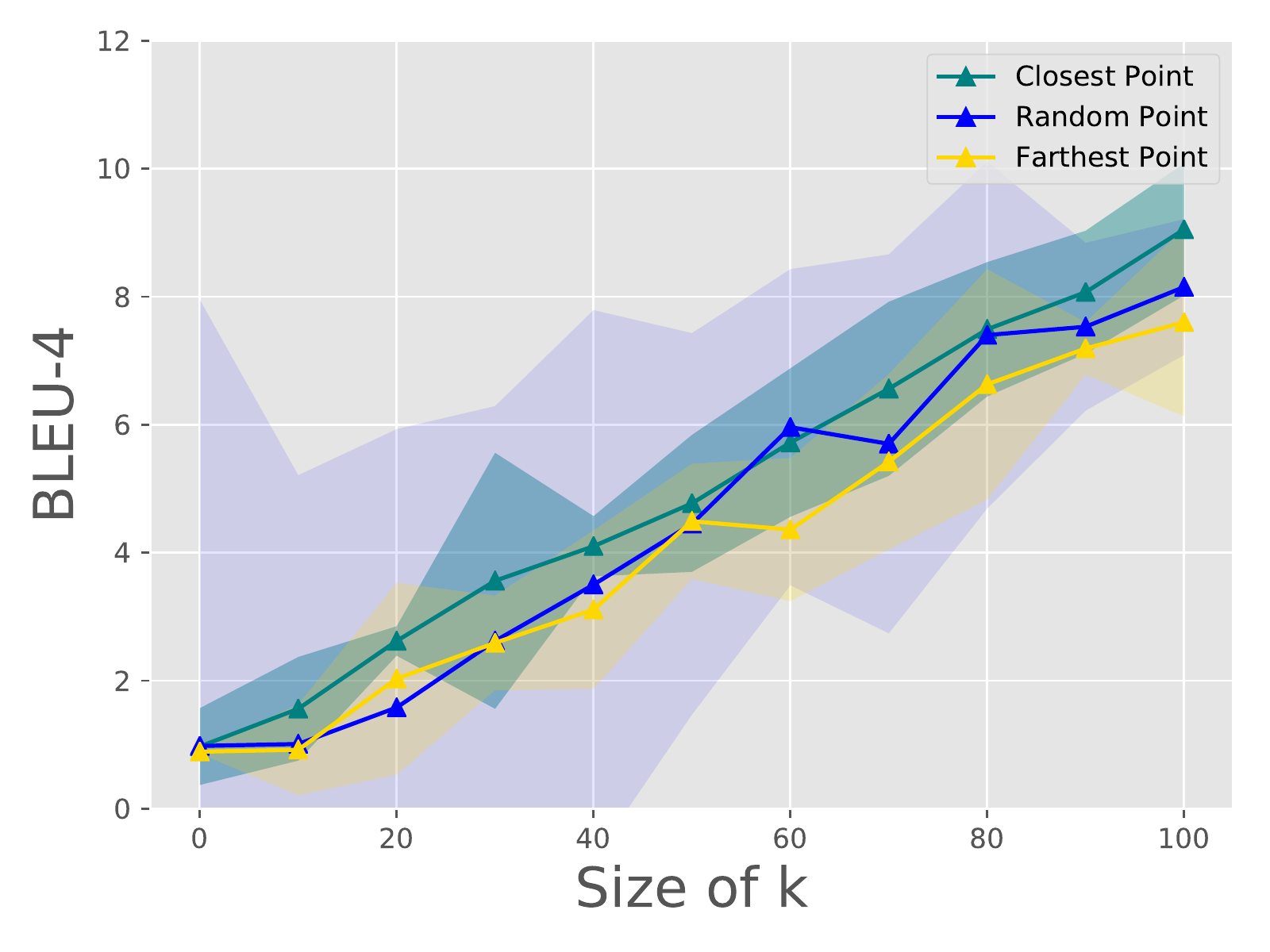}
\caption{ \small
Ablation studies on the SQUAD corpus.
 Performance in BLEU-4 with increasing K between different variants of K-means where selection is based on the \textbf{closest point}, \textbf{Random point}, or \textbf{Farthest point} from the centroid.
}
\label{fig:selection_squad}
\end{figure}
After splitting them into $K$ clusters, we pick from each cluster the data point that is closest to the center. We use the Euclidean distance to select, the same as the metric used for K-means clustering. The intuition is that the test performance usually depends on the nearest neighbor in the training set~\cite{khandelwal2019generalization,rajani2020explaining}. Ideally data points closest to the cluster centers are most representative samples, selecting them will maximize the chance that a similar sample will be found in the training dataset.

\section{Experiments}
We perform our experiments on the following three representative datasets which cover three different text generation tasks:
\begin{enumerate}
    \item Data-to-text: We use the dataset for the E2E challenge~\cite{novikova2017e2e} which contain 50,602 data-text pairs with 8 unique slots in the restaurant domain.
    \item Document Summarization: We use the CNN/Dailymail dataset (non-anonymized version)~\cite{hermann2015teaching} which contains 312,084 document-summary pairs.
    \item Question generation: We use the SQuAD dataset~\cite{rajpurkar2016squad} with over 100k questions. Following \citet{du2017learning}, we focus on the answer-independent scenario to directly generate questions from passages. 
\end{enumerate}

\begin{table}[t]
\resizebox{\columnwidth}{!}{%
\begin{tabular}{c|c|c|c|c|c|c}
\multirow{2}{*}{Embedding} & \multicolumn{2}{c|}{E2E} & \multicolumn{2}{c|}{CNNDM} & \multicolumn{2}{c}{SQUAD} \\ \cline{2-7} 
                           & Mean        & Sum        & Mean         & Sum         & Mean         & Sum         \\ \hline
BART                       & 26.28    & 25.59      & 34.30        & \textbf{34.46}       & 8.89         & 8.56        \\ \hline
BART-FT                    & 26.46    & \textbf{26.32}      & \textbf{36.31}        & 34.18       & \textbf{9.55}              & 8.12        \\ \hline
GloVe                      & 25.18    & 23.36      & 33.59        & {31.45}       & 7.99         & 7.56        \\ \hline
FastText                   & \textbf{27.13}       & 24.85      & 33.23        & 34.30       & 9.33         & \textbf{9.42}        \\ 
\end{tabular}}
\caption{
\small 
 Finetuned BART generation performance comparison on \textbf{E2E}, \textbf{CNNDM}, and \textbf{SQUAD} for various embedding options for the \emph{k-means selection} with k=100. 
}
\label{tab:embedding}
\end{table}
For all experiments, we finetune the open-sourced Bart model~\cite{lewis-etal-2020-bart} as our generative model. Bart is pretrained with a denoising autoencoder objective on large amount of text data and has been the state-of-the-arts for many text generation tasks. To extract vectors used for clustering, we finetune the Bart model with its original self-supervised objective on the unlabeled data, then apply mean pooling over the last hidden states of the encoder.

In the later sections, we will first compare the model performance based on our proposed selection strategy and random sampling, then analyze the variance of them. Finally, we perform an ablation study to see the effects of in-cluster selection and embedding choices.

\paragraph{Comparison of Selection Strategies.}
In Table~\ref{tb:benchmarks}, we compare the model performance based on different selection strategies. Apart from random sampling and our proposed method, we also compare with a lower bound where all instances are randomly sampled from one cluster (within-cluster random). Adding this for comparison aims to illustrate that it is important to select diverse samples across different clusters. The performance scores are averaged over 10 different trials for each selection strategy. As can be seen, K-means based selections consistently outperforms the others. Within-cluster random sampling performs the worst, proving the importance of having diverse samples in the training instance. However, it is worth noting that although random sampling underperforms K-means selection on average, \emph{its upper bound is much higher, suggesting the proposed K-means selection is by no means optimal}. There is still much room for improvement.

\paragraph{Variance of Model Performance.}
Table~\ref{tb:benchmarks} also shows the variance of model performance with different selection strategies. The variance is computed based on 10 different runs. For within-cluster random sampling, the variance comes from both the choice of the cluster and the in-cluster sampling. For K-means selection, the variance comes from the choice of initial center points. We can see random sampling and within-cluster random sampling have a very large variance of up to $7.12$ for $K=100$. This further suggests that comparing few-shot models based on random sampling can be be prone to variability and prevent drawing reliable conclusions. K-means-based selection, on the contrary, is rather robust with random seeds. Therefore, for future work on few-shot text generation, we suggest that models be tested on instances selected from our proposed strategy for a fair comparison.

\paragraph{Effects of In-cluster Selection.}
In Figure~\ref{fig:selection_squad}, we show the effects of the in-cluster selection method. In our proposed method, within each cluster, we select one data point that is closest to the cluster center. To see whether it is important to select the closest data point, we compare with two selection variants that within each cluster, we select (1) one data point randomly sampled from the cluster, and (2) one data point that is farthest to the cluster center. We can observe that the choice of selection does have a big impact on the model performance. Choosing data points farthest to the cluster centers leads to the worst performance. This is consistent with previous findings~\cite{kaushal2018learning,birodkar2019semantic} that data points farthest from cluster centers are usually outliers and less representative. Selecting them might mislead the model to capture non-generic patterns and thereby generalize poorly. In contrast, choosing data points closest to cluster centers performs slightly better than random selection. However, random selection has a much larger variance compared with closest/farthest point selection (shown as shadow).

\paragraph{Effects of Embedding Methods.}
As the K-means clustering is performed on top of the embedding vectors of unlabeled data, the choice of embedding methods could affect the performance on selected points. 
In Table~\ref{tab:embedding}, we show the effects of the different embedding methods. Apart from the finetuned Bart, we compare with embeddings extracted from (1) Bart without being finetuned on the task-specific data, (2) Glove~\cite{pennington2014glove} and (3) FastText~\cite{bojanowski2017enriching}, both finetuned on the task-specific data. For each embedding method, we compare using mean pooling and sum pooling to extract the final vector representation. The results show that finetuned Bart generally outperforms the other embedding choices. 
We attribute this to the similarity in the embedding space between selection with BART embeddings and and the BART generation model.
Moreover, \emph{FastText} offers a strong baseline as it does relatively well on two scenarios in E2E and SQUAD respectively. 
Further, we observe that \emph{mean} pooling is generally better than the \emph{sum} of word vectors, which is also observed in~\citet{chen2018enhancing}.

\paragraph{Human Evaluation.}
To obtain further insights with respect to the generation outputs, five annotators were instructed to evaluate $100$ samples for each of the three tasks to judge (1) whether the text is \emph{fluent} (score 0-5 with 5 being fully fluent), and (2) whether it contains relevant information about its input source (\emph{adequacy}). 
These scores are averaged and presented in Table~\ref{tab:humans}.
For \textbf{Random} selection, we sampled 10 outputs from each of the 10 trials to make it 100 samples, and the same goes for \textbf{IC-random}.
We observe that the K-means algorithm select better subsets of the training samples that allow for better generalizability to unseen input sources. In particular, the outputs are generally more \emph{adequate}. 
However, we see that the \emph{fluency} of outputs remain relatively similar. 

\begin{table}[]
\centering
\resizebox{0.95\columnwidth}{!}{%
\begin{tabular}{c|c|c|c}
                 & E2E & CNNDM & SQUAD \\ \hline
Random           &   4.08/4.15  &  \textbf{4.55}/3.27     &   \textbf{4.62}/{3.84}    \\ \hline
IC-Random &   4.32/3.54  &  3.62/3.01     &   4.23/2.74    \\ \hline
K-means          &   \textbf{4.12}/\textbf{4.24}  &  4.32/\textbf{3.66}     &   4.51/\textbf{3.98}    \\ 
\end{tabular}}
\caption{
\small 
 Human evaluation on 100 samples of the finetuned BART generation performance comparison on \textbf{E2E}, \textbf{CNNDM}, and \textbf{SQUAD}. 
 Scores are presented as (\emph{fluency} / \emph{adequacy}).
}
\label{tab:humans}
\end{table}

\section{Conclusion}
In this work, we target at the unexplored problem of training instance selection for few-shot text generation. We show that random sampling can lead to large variance and suboptimal performance. To address this problem, we propose a selection strategy based on K-mean clustering and demonstrate that it consistently outperforms random sampling, and has much lower variance. 
We further perform a set of ablation studies to analyze the effects of data size, embedding and selection methods, showing that this is still much room for improvement. Future work can consider other clustering methods.

\section*{Acknowledgements}
This research was funded in part by the German Research Foundation (DFG) as part of SFB 248 ``Foundations of Perspicuous Software Systems''. We sincerely thank the anonymous reviewers for their insightful comments that helped us to improve this paper. 

\bibliography{anthology,custom}

\begin{thebibliography}{34}
\expandafter\ifx\csname natexlab\endcsname\relax\def\natexlab#1{#1}\fi

\bibitem[{Agarwal et~al.(2005)Agarwal, Har-Peled, Varadarajan
  et~al.}]{agarwal2005geometric}
Pankaj~K Agarwal, Sariel Har-Peled, Kasturi~R Varadarajan, et~al. 2005.
\newblock Geometric approximation via coresets.
\newblock \emph{Combinatorial and computational geometry}, 52:1--30.

\bibitem[{Arthur and Vassilvitskii(2007)}]{arthur2007k}
David Arthur and Sergei Vassilvitskii. 2007.
\newblock k-means++ the advantages of careful seeding.
\newblock In \emph{Proceedings of the eighteenth annual ACM-SIAM symposium on
  Discrete algorithms}, pages 1027--1035.

\bibitem[{Balcan et~al.(2007)Balcan, Broder, and Zhang}]{balcan2007margin}
Maria-Florina Balcan, Andrei Broder, and Tong Zhang. 2007.
\newblock Margin based active learning.
\newblock In \emph{International Conference on Computational Learning Theory},
  pages 35--50. Springer.

\bibitem[{Birodkar et~al.(2019)Birodkar, Mobahi, and
  Bengio}]{birodkar2019semantic}
Vighnesh Birodkar, Hossein Mobahi, and Samy Bengio. 2019.
\newblock Semantic redundancies in image-classification datasets: The 10\% you
  don't need.
\newblock \emph{arXiv preprint arXiv:1901.11409}.

\bibitem[{Bojanowski et~al.(2017)Bojanowski, Grave, Joulin, and
  Mikolov}]{bojanowski2017enriching}
Piotr Bojanowski, Edouard Grave, Armand Joulin, and Tomas Mikolov. 2017.
\newblock Enriching word vectors with subword information.
\newblock \emph{Transactions of the Association for Computational Linguistics},
  5:135--146.

\bibitem[{Chang et~al.(2020)Chang, Caplinger, Marin, Shen, and
  Demberg}]{chang2020dart}
Ernie Chang, Jeriah Caplinger, Alex Marin, Xiaoyu Shen, and Vera Demberg. 2020.
\newblock Dart: A lightweight quality-suggestive data-to-text annotation tool.
\newblock In \emph{Proceedings of the 28th International Conference on
  Computational Linguistics: System Demonstrations}, pages 12--17.

\bibitem[{Chang et~al.(2021{\natexlab{a}})Chang, Demberg, and
  Marin}]{chang2021jointly}
Ernie Chang, Vera Demberg, and Alex Marin. 2021{\natexlab{a}}.
\newblock Jointly improving language understanding and generation with
  quality-weighted weak supervision of automatic labeling.
\newblock \emph{Proceedings of EACL 2021}.

\bibitem[{Chang et~al.(2021{\natexlab{b}})Chang, Shen, Zhu, Demberg, and
  Su}]{chang2021neural}
Ernie Chang, Xiaoyu Shen, Dawei Zhu, Vera Demberg, and Hui Su.
  2021{\natexlab{b}}.
\newblock Neural data-to-text generation with lm-based text augmentation.
\newblock In \emph{Proceedings of the 16th Conference of the European Chapter
  of the Association for Computational Linguistics: Main Volume}, pages
  758--768.

\bibitem[{Chang et~al.(2021{\natexlab{c}})Chang, Yeh, and
  Demberg}]{chang2021does}
Ernie Chang, Hui-Syuan Yeh, and Vera Demberg. 2021{\natexlab{c}}.
\newblock Does the order of training samples matter? improving neural
  data-to-text generation with curriculum learning.
\newblock \emph{Proceedings of EACL 2021}.

\bibitem[{Chen et~al.(2018)Chen, Ling, and Zhu}]{chen2018enhancing}
Qian Chen, Zhen-Hua Ling, and Xiaodan Zhu. 2018.
\newblock Enhancing sentence embedding with generalized pooling.
\newblock In \emph{Proceedings of the 27th International Conference on
  Computational Linguistics}, pages 1815--1826.

\bibitem[{Chen et~al.(2020)Chen, Eavani, Chen, Liu, and Wang}]{chen2019few}
Zhiyu Chen, Harini Eavani, Wenhu Chen, Yinyin Liu, and William~Yang Wang. 2020.
\newblock Few-shot nlg with pre-trained language model.
\newblock \emph{ACL}.

\bibitem[{Dempster et~al.(1977)Dempster, Laird, and
  Rubin}]{dempster1977maximum}
Arthur~P Dempster, Nan~M Laird, and Donald~B Rubin. 1977.
\newblock Maximum likelihood from incomplete data via the em algorithm.
\newblock \emph{Journal of the Royal Statistical Society: Series B
  (Methodological)}, 39(1):1--22.

\bibitem[{Du et~al.(2017)Du, Shao, and Cardie}]{du2017learning}
Xinya Du, Junru Shao, and Claire Cardie. 2017.
\newblock Learning to ask: Neural question generation for reading
  comprehension.
\newblock In \emph{Association for Computational Linguistics (ACL)}.

\bibitem[{Hermann et~al.(2015)Hermann, Kocisky, Grefenstette, Espeholt, Kay,
  Suleyman, and Blunsom}]{hermann2015teaching}
Karl~Moritz Hermann, Tomas Kocisky, Edward Grefenstette, Lasse Espeholt, Will
  Kay, Mustafa Suleyman, and Phil Blunsom. 2015.
\newblock Teaching machines to read and comprehend.
\newblock In \emph{Advances in neural information processing systems}, pages
  1693--1701.

\bibitem[{Kale(2020)}]{kale2020text}
Mihir Kale. 2020.
\newblock Text-to-text pre-training for data-to-text tasks.
\newblock \emph{arXiv preprint arXiv:2005.10433}.

\bibitem[{Kanungo et~al.(2002)Kanungo, Mount, Netanyahu, Piatko, Silverman, and
  Wu}]{kanungo2002efficient}
Tapas Kanungo, David~M Mount, Nathan~S Netanyahu, Christine~D Piatko, Ruth
  Silverman, and Angela~Y Wu. 2002.
\newblock An efficient k-means clustering algorithm: Analysis and
  implementation.
\newblock \emph{IEEE transactions on pattern analysis and machine
  intelligence}, 24(7):881--892.

\bibitem[{Kaushal et~al.(2018)Kaushal, Sahoo, Doctor, Raju, Shetty, Singh,
  Iyer, and Ramakrishnan}]{kaushal2018learning}
Vishal Kaushal, Anurag Sahoo, Khoshrav Doctor, Narasimha Raju, Suyash Shetty,
  Pankaj Singh, Rishabh Iyer, and Ganesh Ramakrishnan. 2018.
\newblock Learning from less data: Diversified subset selection and active
  learning in image classification tasks.
\newblock \emph{arXiv preprint arXiv:1805.11191}.

\bibitem[{Khandelwal et~al.(2020)Khandelwal, Levy, Jurafsky, Zettlemoyer, and
  Lewis}]{khandelwal2019generalization}
Urvashi Khandelwal, Omer Levy, Dan Jurafsky, Luke Zettlemoyer, and Mike Lewis.
  2020.
\newblock Generalization through memorization: Nearest neighbor language
  models.
\newblock \emph{ICLR}.

\bibitem[{Konyushkova et~al.(2017)Konyushkova, Raphael, and
  Fua}]{konyushkova2017learning}
Ksenia Konyushkova, Sznitman Raphael, and Pascal Fua. 2017.
\newblock Learning active learning from data.
\newblock In \emph{Proceedings of the 31st International Conference on Neural
  Information Processing Systems}, pages 4228--4238.

\bibitem[{Krishna and Murty(1999)}]{krishna1999genetic}
K~Krishna and M~Narasimha Murty. 1999.
\newblock Genetic k-means algorithm.
\newblock \emph{IEEE Transactions on Systems, Man, and Cybernetics, Part B
  (Cybernetics)}, 29(3):433--439.

\bibitem[{Lewis et~al.(2020)Lewis, Liu, Goyal, Ghazvininejad, Mohamed, Levy,
  Stoyanov, and Zettlemoyer}]{lewis-etal-2020-bart}
Mike Lewis, Yinhan Liu, Naman Goyal, Marjan Ghazvininejad, Abdelrahman Mohamed,
  Omer Levy, Veselin Stoyanov, and Luke Zettlemoyer. 2020.
\newblock \href {https://doi.org/10.18653/v1/2020.acl-main.703} {{BART}:
  Denoising sequence-to-sequence pre-training for natural language generation,
  translation, and comprehension}.
\newblock In \emph{Proceedings of the 58th Annual Meeting of the Association
  for Computational Linguistics}, pages 7871--7880, Online. Association for
  Computational Linguistics.

\bibitem[{Li and Liang(2021)}]{li2021prefix}
Xiang~Lisa Li and Percy Liang. 2021.
\newblock Prefix-tuning: Optimizing continuous prompts for generation.
\newblock \emph{arXiv preprint arXiv:2101.00190}.

\bibitem[{Novikova et~al.(2017)Novikova, Du{\v{s}}ek, and
  Rieser}]{novikova2017e2e}
Jekaterina Novikova, Ond{\v{r}}ej Du{\v{s}}ek, and Verena Rieser. 2017.
\newblock The e2e dataset: New challenges for end-to-end generation.
\newblock \emph{arXiv preprint arXiv:1706.09254}.

\bibitem[{Pennington et~al.(2014)Pennington, Socher, and
  Manning}]{pennington2014glove}
Jeffrey Pennington, Richard Socher, and Christopher~D Manning. 2014.
\newblock Glove: Global vectors for word representation.
\newblock In \emph{Proceedings of the 2014 conference on empirical methods in
  natural language processing (EMNLP)}, pages 1532--1543.

\bibitem[{Radford et~al.(2019)Radford, Wu, Child, Luan, Amodei, and
  Sutskever}]{radford2019language}
Alec Radford, Jeffrey Wu, Rewon Child, David Luan, Dario Amodei, and Ilya
  Sutskever. 2019.
\newblock Language models are unsupervised multitask learners.
\newblock \emph{OpenAI blog}, 1(8):9.

\bibitem[{Rajani et~al.(2020)Rajani, Krause, Yin, Niu, Socher, and
  Xiong}]{rajani2020explaining}
Nazneen~Fatema Rajani, Ben Krause, Wengpeng Yin, Tong Niu, Richard Socher, and
  Caiming Xiong. 2020.
\newblock Explaining and improving model behavior with k nearest neighbor
  representations.
\newblock \emph{arXiv preprint arXiv:2010.09030}.

\bibitem[{Rajpurkar et~al.(2016)Rajpurkar, Zhang, Lopyrev, and
  Liang}]{rajpurkar2016squad}
Pranav Rajpurkar, Jian Zhang, Konstantin Lopyrev, and Percy Liang. 2016.
\newblock Squad: 100,000+ questions for machine comprehension of text.
\newblock In \emph{Proceedings of the 2016 Conference on Empirical Methods in
  Natural Language Processing}, pages 2383--2392.

\bibitem[{Schick and Sch{\"u}tze(2020{\natexlab{a}})}]{schick2020few}
Timo Schick and Hinrich Sch{\"u}tze. 2020{\natexlab{a}}.
\newblock Few-shot text generation with pattern-exploiting training.
\newblock \emph{arXiv preprint arXiv:2012.11926}.

\bibitem[{Schick and Sch{\"u}tze(2020{\natexlab{b}})}]{schick2020s}
Timo Schick and Hinrich Sch{\"u}tze. 2020{\natexlab{b}}.
\newblock It's not just size that matters: Small language models are also
  few-shot learners.
\newblock \emph{arXiv preprint arXiv:2009.07118}.

\bibitem[{Settles(2009)}]{settles2009active}
Burr Settles. 2009.
\newblock Active learning literature survey.

\bibitem[{Su et~al.(2020)Su, Shen, Xiao, Zhang, Chang, Zhang, Niu, and
  Zhou}]{su2020moviechats}
Hui Su, Xiaoyu Shen, Zhou Xiao, Zheng Zhang, Ernie Chang, Cheng Zhang, Cheng
  Niu, and Jie Zhou. 2020.
\newblock Moviechats: Chat like humans in a closed domain.
\newblock In \emph{Proceedings of EMNLP 2020}, pages 6605--6619.

\bibitem[{Tur et~al.(2005)Tur, Hakkani-T{\"u}r, and
  Schapire}]{tur2005combining}
Gokhan Tur, Dilek Hakkani-T{\"u}r, and Robert~E Schapire. 2005.
\newblock Combining active and semi-supervised learning for spoken language
  understanding.
\newblock \emph{Speech Communication}, 45(2):171--186.

\bibitem[{Wei et~al.(2015)Wei, Iyer, and Bilmes}]{wei2015submodularity}
Kai Wei, Rishabh Iyer, and Jeff Bilmes. 2015.
\newblock Submodularity in data subset selection and active learning.
\newblock In \emph{International Conference on Machine Learning}, pages
  1954--1963.

\bibitem[{Zhang et~al.(2020)Zhang, Zhao, Saleh, and Liu}]{zhang2020pegasus}
Jingqing Zhang, Yao Zhao, Mohammad Saleh, and Peter Liu. 2020.
\newblock Pegasus: Pre-training with extracted gap-sentences for abstractive
  summarization.
\newblock In \emph{International Conference on Machine Learning}, pages
  11328--11339. PMLR.

\end{thebibliography}
\bibliographystyle{acl_natbib}

\end{document}